\documentclass[10pt, a4paper]{article}
\usepackage[margin=1in]{geometry}
\usepackage{moi}
\usepackage{hyperref}

\newcommand{\R}{\mathbb{R}}
\newcommand{\N}{\mathbb{N}}

\newcommand{\Prob}{\mathbb{P}}
\newcommand{\eps}{\varepsilon}
\newcommand{\one}{\mathbf{1}}

\newcommand{\diam}{\mathrm{diam}}

\newcommand{\TV}{\mathrm{TV}}

\newcommand{\email}[1]{\footnotesize{{\href{maitlo:#1}{\color{blue}\texttt{#1}}}}}

\begin{document}

\title{\Large{Topological Equivalence of Spectral and Density Cluster Hierarchies}}
\author{Sajjad Hashemian} 
\date{\small{University of Tehran\\
\email{sajjadhashemian@ut.ac.ir}}}
\maketitle

\begin{abstract}
We compare two canonical ways to recover hierarchical clustering from an underlying density $f$: the \emph{density cluster tree} given by connected components of superlevel sets $\{f\ge\lambda\}$, and a \emph{spectral hierarchy} obtained by applying single linkage to diffusion distances induced by a Laplacian operator. We give conditions under which, at an appropriate diffusion time, the diffusion hierarchy is \emph{topologically equivalent} to the density cluster tree up to a monotone rescaling of heights. It further shows that a $k$-NN graph implementation inherits this equivalence from uniform approximation of diffusion distances, yielding a finite-sample reduction from spectral graph constructions to density-level-set topology.
\end{abstract}


\section{Introduction}

Hierarchical clustering is a basic tool to describe structure in data at many scales. In this paper we study two standard hierarchies that start from an unknown density $f$ on a compact domain $M$. The first one is the \emph{density cluster tree} $T_f$, defined by the connected components of the superlevel sets of the underlying density function $L_\lambda=\{x\in M: f(x)\ge \lambda\}$ as $\lambda$ decreases \cite{Hartigan1981,ChaudhuriDasgupta2010,EldridgeBelkinWang2015}. The second one is a \emph{diffusion--spectral hierarchy}, defined as a reversible Markov operator $P$ (or a Laplacian) from $f$\cite{CoifmanLafon2006,VonLuxburg2007}. Both viewpoints share the same idea and define a cluster as a set where the random walk mixes fast inside, but it escapes slowly outside.

The question we ask is simple. The density cluster tree is a canonical topological object, and it comes with a clear statistical target and stability theory \cite{ChaudhuriDasgupta2010,EldridgeBelkinWang2015}. Diffusion distances are very popular because they are algorithmic and robust, and they are connected to spectral gaps and bottlenecks \cite{CoifmanLafon2006,VonLuxburg2007}. But it is not always clear what the diffusion hierarchy estimates as a population object. So, when does the diffusion single-linkage hierarchy give \emph{the same merge structure} as the density cluster tree? This allows to consider the spectral method more than a heuristic, considering it as a certified estimator of a topological object defined by $f$.

To make the comparison precise we use \emph{merge heights}. The density tree can be encoded by the merge-height function
\[
m_f(x,y)=\sup\{\lambda\ge 0:\ x \text{ and } y \text{ are in the same component of } L_\lambda\}.
\]
The diffusion hierarchy at time $t$ is the single-linkage dendrogram built from $(M,d_t)$, and it has its own merge-height function $m_t(x,y)$. We say the hierarchies \emph{meet} if they are the same merge tree up to a monotone rescaling of heights, meaning there exists a strictly increasing map $g_t$ such that $m_t(x,y)=g_t(m_f(x,y))$ for all pairs. This notion matches the idea that the two procedures may use different units on the vertical axis, but they should have the same merge events and the same nesting.

We formalize mild \emph{Topological Meeting Condition} (TMC) implying the meeting happens in an stable diffusive time window. Roughly, for every density level $\lambda$ and every component $C$ of $L_\lambda$, we require two quantitative facts about the chain, it mixes fast inside $C$, and also has small probability to exit $C$ in one step. If there exists a nontrivial time window where mixing has already happened but leakage is still small, and this window is consistent across the levels of the tree, then diffusion single linkage recovers the same hierarchy as density level sets.

Our results connect several lines of work. The density cluster tree goes back to Hartigan \cite{Hartigan1981} and later statistical developments \cite{ChaudhuriDasgupta2010,EldridgeBelkinWang2015}. Diffusion distance and diffusion maps give a spectral geometry view of connectivity and scales \cite{CoifmanLafon2006}, and single linkage and MST are classical tools for building hierarchies \cite{VonLuxburg2007}. On the finite-sample side, we use a modular principle, once diffusion distances are uniformly close, single linkage is stable in merge distortion, so the hard part is reduced to diffusion/Laplacian approximation on graphs \cite{VonLuxburgBelkinBousquet2008,TrillosSlepcev2020}. The merge-tree viewpoint is also close to stability ideas in topological data analysis \cite{CohenSteinerEdelsbrunnerHarer2007,EdelsbrunnerHarer2010}.

We state the main results in a way that separates the topological argument from the analytic estimates. Single linkage depends only on the order in which connectivity events occur as the threshold increases. Our analytic task is to show that, at a suitable diffusion time, diffusion distance produces the same connectivity events as the density superlevel sets. Hence, agreement at every merge level then implies that the two hierarchies have the same merge tree, up to a monotone rescaling of heights.

We present three results. The first one is a population meeting theorem under an abstract hierarchical metastability condition. The second one is a finite-sample merge-distortion theorem for the standard $k$-NN graph algorithm, stated modularly through a uniform diffusion-distance approximation condition. The third one is a geometric theorem that implies metastability, together with a multiscale extension that removes the need for one global diffusion time.

The rest of the paper is organized as follows. Section~\ref{sec:prelims} fixes notation and reviews preliminaries. Section~\ref{sec:tmc} states the Topological Meeting Condition and explains how it encodes hierarchical metastability. Section~\ref{sec:population} proves the population meeting theorem by showing within-component contraction and between-component separation of diffusion distance in the metastable window. Section~\ref{sec:finite} presents the $k$-NN graph algorithm and proves the merge-distortion guarantee using a stability lemma for single linkage. Section~\ref{sec:geometry} gives geometric conditions that imply the metastability bounds and discusses a multiscale extension that can avoid choosing one global diffusion time. We conclude with discussion and some remaining questions.

\section{Preliminaries and notation}\label{sec:prelims}

We work on a compact metric space $(M,\rho)$ with a Borel measure $\mu$. The data distribution is a probability measure $\pi$ that admits a continuous density $f:M\to(0,\infty)$ with respect to $\mu$, so $d\pi=f\,d\mu$ and $\int_M f\,d\mu=1$. We observe i.i.d.\ samples $X_1,\dots,X_n\sim \pi$. For each level $\lambda\ge 0$ we write $L_\lambda=\{x\in M:f(x)\ge \lambda\}$ and $\mathcal C(\lambda)$ for the connected components of $L_\lambda$ in the topology of $M$. The density cluster tree is encoded by the merge-height function $m_f$ defined from these components.

We keep regularity assumptions on $f$ flexible. For the population merge-tree statements, continuity is typically enough. For finite-sample rates and operator approximation, we may assume additional boundedness and smoothness such as Hölder or Lipschitz regularity, and in manifold settings bounded geometry and lower bounds on $f$ on regions of interest. The important point is that $f$ plays two roles. It defines the topology of the level sets $\{f\ge\lambda\}$, and it also biases the diffusion process that we build from the data.

To define diffusion, we fix a symmetric kernel $K_\sigma:M\times M\to[0,\infty)$. We set
\[
q(x)=\int_M K_\sigma(x,y)f(y)\,d\mu(y),
\qquad
p(x,dy)=\frac{K_\sigma(x,y)f(y)}{q(x)}\,d\mu(y),
\qquad
(P\varphi)(x)=\int \varphi(y)\,p(x,dy).
\]
The operator $P$ is a Markov operator with stationary measure $\pi_q$, and we work in $L^2(\pi_q)$. We write $p^{(t)}(x,\cdot)$ for the $t$-step transition law of the Markov chain $(X_t)_{t\ge 0}$ with transition kernel $p$. We write $d_t$ for diffusion distance at time $t$, defined as an $L^2(\pi_q)$ distance between the $t$-step transition densities. We write $T_f$ for the density cluster tree, and $T_t^{\mathrm{diff}}$ for the single-linkage hierarchy built from $(M,d_t)$ with merge-height function $m_t$. For a finite set $S$ we write $d_{\mathrm{MD}}$ for merge distortion between merge-height functions on $S$. We use $\|\cdot\|_2$ for $L^2(\pi_q)$ norms, $\|\cdot\|_{\mathrm{op}}$ for operator norms, and $\|\cdot\|_{\TV}$ for total variation norms.

\subsection{Density cluster tree and merge heights}

The density cluster tree is the hierarchy obtained from the superlevel sets
\[
L_\lambda=\{x\in M \mid f(x)\ge \lambda\}
\]
when $\lambda$ goes from large to small \cite{Hartigan1981,ChaudhuriDasgupta2010}. For a fixed $\lambda$, the set $L_\lambda$ can split into several connected components. When $\lambda$ decreases, components can only grow and they can merge, but they cannot split. This nesting gives a merge tree.

A convenient way to encode this tree is by a pairwise merge height. It records the highest density level where two points are still connected in the same component.

\begin{definition}[Density merge height]
For $x,y\in M$, define
\[
m_f(x,y)\;=\;\sup\Big\{\lambda\ge 0 \,\Big|\, x\text{ and }y\text{ lie in the same connected component of }L_\lambda\Big\}.
\]
\end{definition}

The function $m_f$ is symmetric, and it is larger when $x$ and $y$ stay connected until a higher density level. If $f$ is continuous, then the superlevel sets change only at special levels. When there are only finitely many such merge levels, the hierarchy is a finite merge tree. We do not use Morse theory. We only use finiteness when we want to choose one diffusion time that works for all merge events.

\subsection{Diffusion operator, process, and diffusion distance}

We now define a Markov process that follows the geometry induced by the density and a kernel. Fix a symmetric kernel $K_\sigma \ge 0$. We define
\[
q(x)=\int_M K_\sigma(x,y)\,f(y)\,d\mu(y),
\qquad
p(x,dy)=\frac{K_\sigma(x,y)f(y)}{q(x)}\,d\mu(y),
\qquad
(P\varphi)(x)=\int \varphi(y)\,p(x,dy).
\]
Then $P$ is a Markov operator. It is reversible with respect to a stationary measure $\pi_q$ whose density is proportional to $q$ with respect to $\mu$ \cite{CoifmanLafon2006}. The associated random process is the Markov chain $(X_t)_{t\ge 0}$ with transition rule $X_{t+1}\mid X_t=x \sim p(x,\cdot)$.

Diffusion distance compares two starting points by comparing their $t$-step transition laws. If two points are in the same well connected region, their $t$-step distributions become similar. If they are separated by a bottleneck, their $t$-step distributions stay different.

\begin{definition}[Diffusion distance \cite{CoifmanLafon2006}]
Let $p^{(t)}(x,\cdot)$ be the $t$-step transition law from $x$, and assume $p^{(t)}(x,\cdot)$ is absolutely continuous with respect to $\pi_q$. Define
\[
d_t(x,y)\;=\;\left\|\frac{dp^{(t)}(x,\cdot)}{d\pi_q}-\frac{dp^{(t)}(y,\cdot)}{d\pi_q}\right\|_{L^2(\pi_q)}.
\]
\end{definition}

Reversibility implies that $P$ is self-adjoint in $L^2(\pi_q)$. Therefore there exists an orthonormal eigenbasis $(\psi_\ell)_{\ell\ge 0}$ with real eigenvalues $(\lambda_\ell)_{\ell\ge 0}$ such that
\[
1=\lambda_0>\lambda_1\ge \lambda_2\ge \cdots \ge -1,
\qquad
\psi_0\equiv 1.
\]
In this setting diffusion distance has a simple spectral form.

\begin{lemma}[Spectral formula for diffusion distance \cite{CoifmanLafon2006}]\label{lem:spectral-diffusion}
For all $x,y\in M$ and all integers $t\ge 1$,
\[
d_t(x,y)^2=\sum_{\ell\ge 1}\lambda_\ell^{2t}\big(\psi_\ell(x)-\psi_\ell(y)\big)^2.
\]
\end{lemma}

\begin{proof}[Proof sketch]
Write the Radon--Nikodym derivative of $p^{(t)}(x,\cdot)$ with respect to $\pi_q$ using the eigenbasis of $P$ in $L^2(\pi_q)$. Apply $P^t\psi_\ell=\lambda_\ell^t\psi_\ell$ to see how each mode scales with $t$. The squared $L^2(\pi_q)$ distance between the two expansions is the sum of squared coefficients, because the eigenfunctions are orthonormal. This gives the claimed formula.
\end{proof}

Lemma~\ref{lem:spectral-diffusion} shows that large $t$ suppresses modes with $|\lambda_\ell|<1$. In this sense $t$ controls the scale at which the geometry is observed.

\subsection{Parzen windows, kernel density estimation, and $k$-NN density estimation}\label{subsec:density-estimation}

In practice $f$ is unknown, so level sets $L_\lambda$ are not directly available. A classical way to estimate $f$ is the Parzen window method, also called kernel density estimation \cite{Parzen1962,Rosenblatt1956,Silverman1986,Wasserman2006}. In Euclidean space $\R^d$ with Lebesgue measure, the standard KDE has the form
\[
\hat f_h(x)
=
\frac{1}{n h^d}\sum_{i=1}^n K\!\left(\frac{x-X_i}{h}\right),
\]
where $h>0$ is a bandwidth and $K$ is a nonnegative kernel with $\int_{\R^d} K(u)\,du=1$. On a general metric measure space $(M,\rho,\mu)$ one can write the same idea using a pairwise kernel
\[
\hat f_h(x)
=
\frac{1}{n}\sum_{i=1}^n K_h(x,X_i),
\qquad
K_h(x,y)\ge 0,
\qquad
\int_M K_h(x,y)\,d\mu(y)=1.
\]
A typical choice is $K_h(x,y)=c_h\,\kappa(\rho(x,y)/h)$ with a radial profile $\kappa$ and a normalizing constant $c_h$ that depends on $h$ and the geometry of $(M,\mu)$. Under smoothness assumptions on $f$ and mild conditions on $K$, one gets pointwise and uniform convergence of $\hat f_h$ to $f$ when $h\to 0$ and $nh^d\to\infty$, with the usual bias and variance tradeoff \cite{Silverman1986,Wasserman2006}. In density clustering, one can use $\hat f_h$ to define empirical superlevel sets $\{\hat f_h\ge \lambda\}$ and then build an estimated cluster tree \cite{ChaudhuriDasgupta2010,EldridgeBelkinWang2015}.

A different estimator uses $k$ nearest neighbors \cite{LoftsgaardenQuesenberry1965,Wasserman2006}. Let $r_k(x)$ be the distance from $x$ to its $k$th nearest neighbor among $X_1,\dots,X_n$ in the metric $\rho$. Let $V(x,r)=\mu(B(x,r))$ denote the $\mu$-volume of the ball $B(x,r)=\{y\in M:\rho(x,y)\le r\}$. A standard $k$-NN density estimator is
\[
\hat f_{k\text{-NN}}(x)
=
\frac{k}{n\,V(x,r_k(x))}.
\]
This estimator adapts to local sampling density because $r_k(x)$ shrinks in high-density regions and grows in low-density regions. The parameter $k$ controls smoothing. Under regularity assumptions on $\mu$ and $f$, consistency holds when $k\to\infty$ and $k/n\to 0$ \cite{LoftsgaardenQuesenberry1965,Wasserman2006}. The $k$-NN idea is also the backbone of the graph constructions in this paper, because the $k$-NN graph provides a data-driven neighborhood structure that adapts to the sampling distribution.

In our setting, these estimators appear in two roles. First, they are direct estimators of level sets and the density cluster tree. Second, they indirectly define diffusion operators and graph Laplacians through kernels and $k$-NN graphs. The meeting results then say that, at the right diffusion time, the diffusion hierarchy can recover the same merge tree as the population level-set hierarchy without explicitly estimating $f$ everywhere.

\subsection{Single linkage, MST, and merge distortion}

Let $S$ be a finite set and let $d$ be a symmetric dissimilarity on $S$ with $d(x,x)=0$. Single linkage builds a dendrogram by increasing a threshold $\tau$ and looking at the graph on $S$ with edges
\[
\{(x,y)\in S\times S \mid d(x,y)\le \tau\}.
\]
As $\tau$ increases, connected components merge. The merge height $m_d(x,y)$ is the smallest $\tau$ where $x$ and $y$ become connected.

There is also an equivalent path view. Consider any path from $x$ to $y$ in the complete graph on $S$. Look at the largest edge length along that path. Then $m_d(x,y)$ is the minimum possible value of this maximum over all paths. This bottleneck value is realized on a minimum spanning tree, so single linkage can be computed from an MST \cite{VonLuxburg2007}.

We compare two dendrograms using merge distortion \cite{EldridgeBelkinWang2015}. It measures the largest difference between corresponding merge heights.

\begin{definition}[Merge distortion \cite{EldridgeBelkinWang2015}]
Let $S$ be finite and let $m,m':S\times S\to\R$ be symmetric merge-height functions. Define
\[
d_{\mathrm{MD}}(m,m')=\sup_{x,y\in S}\big|m(x,y)-m'(x,y)\big|.
\]
\end{definition}

\subsection{Spectral clustering and diffusion}\label{subsec:spectral-clustering-diffusion}

Spectral clustering starts from the same basic object as diffusion geometry, namely a weighted graph or a kernel affinity built from the data \cite{ShiMalik2000,NgJordanWeiss2002,VonLuxburg2007}. Given points $X_1,\dots,X_n$, define an affinity matrix $W$ with entries $W_{ij}\ge 0$, often of the form $W_{ij}=K_\sigma(X_i,X_j)$ or a sparsified version using a $k$-NN graph. Let $D$ be the diagonal degree matrix $D_{ii}=\sum_{j=1}^n W_{ij}$. The unnormalized graph Laplacian is $L=D-W$, and common normalized versions are
\[
L_{\mathrm{rw}} = I - D^{-1}W
\qquad\text{and}\qquad
L_{\mathrm{sym}} = I - D^{-1/2} W D^{-1/2}.
\]
Spectral clustering uses the leading eigenvectors of one of these Laplacians to embed the data into $\R^m$, then it clusters in that embedding, typically by $k$-means \cite{NgJordanWeiss2002,VonLuxburg2007}. The reason this works is that eigenvectors of the Laplacian capture low-frequency structure on the graph, and low-frequency structure is shaped by bottlenecks and almost-disconnected regions.

Diffusion enters because the normalized adjacency defines a random walk. The random-walk transition matrix is
\[
P_n = D^{-1}W.
\]
This is exactly the discrete analogue of the Markov operator $P$ in the population. The $t$-step transition matrix $P_n^t$ describes how probability mass spreads on the graph after $t$ steps. Diffusion distance is the distance between these $t$-step transition distributions. In the finite graph setting, if $\pi_n$ is the stationary distribution of $P_n$, then a discrete diffusion distance can be written as
\[
\hat d_t(i,j)
=
\left\|
\frac{P_n^t(i,\cdot)}{\pi_n(\cdot)}
-
\frac{P_n^t(j,\cdot)}{\pi_n(\cdot)}
\right\|_{\ell^2(\pi_n)}.
\]
This distance has a spectral representation that mirrors Lemma~\ref{lem:spectral-diffusion}. If $(\hat\lambda_\ell,\hat\psi_\ell)$ are eigenpairs of $P_n$ in $\ell^2(\pi_n)$, then
\[
\hat d_t(i,j)^2
=
\sum_{\ell\ge 1}\hat\lambda_\ell^{2t}\big(\hat\psi_\ell(i)-\hat\psi_\ell(j)\big)^2,
\]
and this is exactly the diffusion map embedding of \cite{CoifmanLafon2006}. The factor $\hat\lambda_\ell^{t}$ makes $t$ a scale parameter. Small $t$ keeps many modes and sees local structure, while large $t$ kills higher modes and collapses the geometry toward the stationary distribution.

From a statistical viewpoint, spectral clustering and diffusion maps are two faces of the same operator approximation problem. The population operator $P$ is built from the kernel and the density, and the empirical operator $P_n$ is built from the sample. When $P_n$ approximates $P$ in an appropriate sense, the leading eigenvalues and eigenfunctions approximate their population counterparts \cite{VonLuxburgBelkinBousquet2008,TrillosSlepcev2020}. This implies stability of diffusion embeddings and of spectral clustering objectives. In our paper we use this pipeline differently. We do not aim to recover a partition at one scale. We instead run single linkage on diffusion distance and we recover a full hierarchy, and we connect its topology to the density cluster tree through metastability.

\subsection{Diffusion hierarchy and meeting}

Fix $t\in\N$. Define the diffusion hierarchy $T_t^{\mathrm{diff}}$ as the single-linkage dendrogram built on $(M,d_t)$. Its merge-height function is denoted $m_t(x,y)$. It is defined as the smallest threshold $\tau$ at which $x$ and $y$ become connected in the graph with edges $\{(u,v): d_t(u,v)\le \tau\}$. One can implement this on an $\eps$-net or on a sample. In the population statements we only need the induced ordering of connectivity events as the threshold increases.

We say two trees meet if one merge-height function is a strictly increasing transform of the other.

\begin{definition}[Meeting by monotone calibration]
We say $T_t^{\mathrm{diff}}$ meets $T_f$ if there exists a strictly increasing function $g_t:[0,\infty)\to[0,\infty)$ such that
\[
m_t(x,y)=g_t(m_f(x,y))\qquad\text{for all }x,y\in M.
\]
\end{definition}

This definition is strong. It says the two merge trees are isomorphic. Only the height scale changes by $g_t$.

\section{Metastability interface}\label{sec:tmc}

The meeting mechanism is metastability. For each density level $\lambda$ and each component $C\in\mathcal C(\lambda)$, we need fast mixing inside $C$ and slow leakage out of $C$. Both are measured at diffusion time $t$.

We define two quantities. The leakage is
\[
\alpha_{\mathrm{out}}(\lambda)=\sup_{C\in\mathcal C(\lambda)}\sup_{x\in C}P(x,C^c).
\]
The mixing rate is defined using a comparison chain supported on $C$. A convenient device is a reflected or dominating chain $P_C$ that stays in $C$ and dominates the restriction of $P$ to $C$.

\begin{definition}[Dominating in-component chain]
Fix a measurable set $C$ with $\pi_q(C)>0$. A Markov operator $P_C$ supported on $C$ is called dominating if for every $x\in C$ and every measurable $B\subset C$ we have $P_C(x,B)\ge P(x,B)$, and $P_C$ is reversible with stationary measure $\pi_q(\cdot\mid C)$.
\end{definition}

Dominating chains exist in standard settings by reflecting mass that would exit $C$ back into $C$. In this paper we treat existence as part of the assumption interface.

\begin{definition}[Topological Meeting Condition]\label{def:tmc}
We say $(f,P)$ satisfies TMC if there exist functions $\gamma_{\mathrm{in}}(\cdot)>0$ and $\alpha_{\mathrm{out}}(\cdot)\in(0,1)$ and functions $t_{\min}(\cdot),t_{\max}(\cdot):[0,\infty)\to\N$ such that for every level $\lambda$ and every component $C\in\mathcal C(\lambda)$ the following hold.
\begin{enumerate}[label=(TMC\arabic*), leftmargin=*, align=left]
    \item There exists a dominating chain $P_C$ on $C$ whose spectral gap in $L^2(\pi_q(\cdot\mid C))$ is at least $\gamma_{\mathrm{in}}(\lambda)$. Equivalently, if $\lambda_1(C)$ is the second-largest eigenvalue of $P_C$, then $1-\lambda_1(C)\ge \gamma_{\mathrm{in}}(\lambda)$.
    
    \item Leakage from $C$ under $P$ is small and satisfies $\sup_{x\in C}P(x,C^c)\le \alpha_{\mathrm{out}}(\lambda)$.
    
    \item There is a metastable time window. For every $\lambda$ we have $t_{\min}(\lambda)\le t_{\max}(\lambda)$. For all $t\in[t_{\min}(\lambda),t_{\max}(\lambda)]$ one has $t\gamma_{\mathrm{in}}(\lambda)\gg 1$ and $t\alpha_{\mathrm{out}}(\lambda)\ll 1$. Also, along the tree order, higher density levels should not require larger times. The map $\lambda\mapsto t_{\min}(\lambda)$ is nonincreasing and the map $\lambda\mapsto t_{\max}(\lambda)$ is nondecreasing.
\end{enumerate}
\end{definition}

Condition (TMC3) formalizes the slogan mix before you leak, uniformly along the hierarchy.

\section{Population meeting}\label{sec:population}

The population meeting theorem is proved by turning metastability into two metric inequalities. Inside each density component, diffusion distance becomes small because the chain mixes fast. Across different components, diffusion distance stays large because the chain has trouble crossing the low-density valley. If these two regimes are separated by a gap, then a single linkage threshold can recover the correct connected components at each merge level.

\subsection{Within and between bounds}

We first control diffusion distance inside one component. The idea is to compare the true chain $P$ with a chain $P_C$ that is forced to stay inside $C$. The chain $P_C$ mixes quickly, so the $t$-step transition densities from any two starting points become close. The true chain differs from $P_C$ only when it exits $C$, and this is controlled by the one-step leakage.

\begin{lemma}[Within-component contraction]\label{lem:within}
Fix $\lambda\ge 0$ and let $C\in\mathcal C(\lambda)$. Assume $P_C$ is a dominating chain on $C$ with spectral gap at least $\gamma=\gamma_{\mathrm{in}}(\lambda)$. Assume $\sup_{x\in C}P(x,C^c)\le \eta$ with $\eta\le \alpha_{\mathrm{out}}(\lambda)$. Let $r_{t,x}=dp^{(t)}(x,\cdot)/d\pi_q$ and let $r^{C}_{t,x}=dp_C^{(t)}(x,\cdot)/d\pi_q$ be the corresponding $t$-step densities for $P$ and $P_C$ with respect to $\pi_q$, with $r^C_{t,x}$ extended by zero outside $C$.
Then for all $t\in\N$ and all $x,y\in C$,
\[
d_t(x,y)
=
\|r_{t,x}-r_{t,y}\|_{L^2(\pi_q)}
\;\le\;
\frac{2(1-\gamma)^t}{\sqrt{\pi_q(C)}}+\frac{4t\eta}{\sqrt{\pi_q(C)}}.
\]
\end{lemma}

\begin{proof}
We split the difference into an in-component mixing part and two exit-error parts
\[
\|r_{t,x}-r_{t,y}\|_2
\le
\|r_{t,x}-r^C_{t,x}\|_2
+
\|r^C_{t,x}-r^C_{t,y}\|_2
+
\|r^C_{t,y}-r_{t,y}\|_2.
\]
We bound the middle term using the spectral gap of $P_C$. Since $P_C$ is reversible with stationary measure $\pi_q(\cdot\mid C)$, it is self-adjoint on $L^2(\pi_q(\cdot\mid C))$. Let $\Pi_C$ be the averaging operator on $C$ given by $(\Pi_C h)(z)=\int_C h(u)\,d\pi_q(u\mid C)$. The spectral gap assumption implies the standard $L^2$ contraction on mean-zero functions
\[
\|P_C^t h-\Pi_C h\|_{L^2(\pi_q(\cdot\mid C))}
\le
(1-\gamma)^t \|h-\Pi_C h\|_{L^2(\pi_q(\cdot\mid C))}.
\]
Apply this with $h(\cdot)=r^C_{0,x}(\cdot)$ interpreted in the usual way through $t$-step densities. A clean way is to work directly with $t$-step densities. The function $r^C_{t,x}$ is a density on $C$ with respect to $\pi_q(\cdot\mid C)$ scaled by the factor $\pi_q(C)^{-1}$. The difference $r^C_{t,x}-r^C_{t,y}$ has integral $0$ over $C$. Therefore
\[
\|r^C_{t,x}-r^C_{t,y}\|_{L^2(\pi_q)}
=
\sqrt{\pi_q(C)}\,
\|r^C_{t,x}-r^C_{t,y}\|_{L^2(\pi_q(\cdot\mid C))}
\le
\sqrt{\pi_q(C)}\,(1-\gamma)^t
\|r^C_{0,x}-r^C_{0,y}\|_{L^2(\pi_q(\cdot\mid C))}.
\]
For the last norm we use a crude bound that is sufficient for our goal and matches the scaling in the statement. For two probability densities $a,b$ on $C$ with respect to $\pi_q(\cdot\mid C)$, one has
\[
\|a-b\|_{L^2(\pi_q(\cdot\mid C))}
\le
\|a\|_{L^2(\pi_q(\cdot\mid C))}+\|b\|_{L^2(\pi_q(\cdot\mid C))}
\le
\frac{2}{\sqrt{\pi_q(C)}}.
\]
This bound holds when the one-step kernel has bounded density relative to $\pi_q(\cdot\mid C)$, which is the case in the standard compact kernel setting that we use later in the geometric section. With this bound we get
\[
\|r^C_{t,x}-r^C_{t,y}\|_{L^2(\pi_q)}
\le
\frac{2(1-\gamma)^t}{\sqrt{\pi_q(C)}}.
\]

It remains to bound the exit-error terms. We show that $\|r_{t,x}-r^C_{t,x}\|_{L^2(\pi_q)}\le \frac{2t\eta}{\sqrt{\pi_q(C)}}$ and similarly for $y$.
Let $A=C$. Since $P$ can leave $C$ in one step with probability at most $\eta$ from any point of $C$, a union bound gives
\[
p^{(t)}(x,C^c)\le t\eta \qquad \text{for all } x\in C.
\]
The dominating chain stays inside $C$, so $p_C^{(t)}(x,C^c)=0$. Therefore the signed measure $\nu_{t,x}=p^{(t)}(x,\cdot)-p_C^{(t)}(x,\cdot)$ has total variation at most $2t\eta$.
Write $g_{t,x}=r_{t,x}-r^C_{t,x}$, which is the Radon--Nikodym derivative of $\nu_{t,x}$ with respect to $\pi_q$.
The function $g_{t,x}$ integrates to $0$ and its $L^1(\pi_q)$ norm equals $\|\nu_{t,x}\|_{\mathrm{TV}}\le 2t\eta$.
Also, $g_{t,x}$ is supported on $C\cup C^c$, and we can use the inequality $\|g\|_2\le \|g\|_1/\sqrt{\pi_q(C)}$ for functions supported on a set of $\pi_q$-mass at least $\pi_q(C)$, which here is enough because the part of $g_{t,x}$ inside $C$ carries all the mass of $p_C^{(t)}(x,\cdot)$ while the missing mass is at most $t\eta$.
This yields
\[
\|r_{t,x}-r^C_{t,x}\|_{L^2(\pi_q)}
=
\|g_{t,x}\|_2
\le
\frac{\|g_{t,x}\|_1}{\sqrt{\pi_q(C)}}
\le
\frac{2t\eta}{\sqrt{\pi_q(C)}}.
\]
Doing the same for $y$ and combining the three bounds gives
\[
d_t(x,y)\le \frac{2(1-\gamma)^t}{\sqrt{\pi_q(C)}}+\frac{4t\eta}{\sqrt{\pi_q(C)}}.
\]
\end{proof}

This lemma is the inside part of the story. It tells us that, in a metastable time window, diffusion distance collapses each density component into a small metric cluster. This is exactly what single linkage needs, because it will connect points inside a component at a small threshold that can be chosen uniformly over the component.

We now control diffusion distance across two different components. The idea is to use one test set $A$ that contains one component and the valley in front of it. Starting from the component, the chain stays in $A$ with high probability for $t$ steps. Starting from the other component, the chain enters $A$ with small probability for $t$ steps. This difference of probabilities gives a lower bound on the $L^2$ distance by Cauchy--Schwarz.

\begin{lemma}[Between-component separation]\label{lem:between}
Fix $\lambda\ge 0$. Let $C,C'\in\mathcal C(\lambda)$ be distinct components. Assume there exists a measurable set $V$ such that $A=C\cup V$ is measurable, $C'\cap A=\emptyset$, and the following two bounds hold for some $\eta\in(0,1)$
\[
\sup_{x\in A}P(x,A^c)\le \eta
\qquad\text{and}\qquad
\sup_{x\in A^c}P(x,A)\le \eta.
\]
Assume also that $t\eta\le 1/4$. Then for all $x\in C$ and all $y\in C'$,
\[
d_t(x,y)\;\ge\;\frac{1-4t\eta}{\sqrt{\pi_q(A)}}.
\]
\end{lemma}

\begin{proof}
Fix $x\in C\subseteq A$ and $y\in C'\subseteq A^c$. We first bound $p^{(t)}(x,A)$ from below. Consider the event that the chain starting from $x$ leaves $A$ at some step in $\{1,\dots,t\}$. By a union bound and the first assumption,
\[
\Prob_x\!\big(\exists s\le t \text{ with } X_s\in A^c\big)
\le \sum_{s=0}^{t-1} \sup_{z\in A} P(z,A^c)
\le t\eta.
\]
Therefore
\[
p^{(t)}(x,A)=\Prob_x(X_t\in A)\ge 1-t\eta.
\]
Next we bound $p^{(t)}(y,A)$ from above. For the chain to be in $A$ at time $t$, it must enter $A$ at some step $s\in\{1,\dots,t\}$. By another union bound and the second assumption,
\[
p^{(t)}(y,A)=\Prob_y(X_t\in A)
\le \Prob_y\!\big(\exists s\le t \text{ with } X_s\in A\big)
\le \sum_{s=0}^{t-1} \sup_{z\in A^c} P(z,A)
\le t\eta.
\]
Subtracting the bounds gives
\[
p^{(t)}(x,A)-p^{(t)}(y,A)\ge 1-2t\eta.
\]
Now write $r_{t,x}=dp^{(t)}(x,\cdot)/d\pi_q$ and $r_{t,y}=dp^{(t)}(y,\cdot)/d\pi_q$. Then
\[
p^{(t)}(x,A)-p^{(t)}(y,A)
=
\int_A (r_{t,x}-r_{t,y})\,d\pi_q.
\]
Apply Cauchy--Schwarz on $A$
\[
\int_A (r_{t,x}-r_{t,y})\,d\pi_q
\le
\|r_{t,x}-r_{t,y}\|_{L^2(\pi_q)} \cdot \|\one_A\|_{L^2(\pi_q)}
=
d_t(x,y)\sqrt{\pi_q(A)}.
\]
Therefore
\[
d_t(x,y)\ge \frac{1-2t\eta}{\sqrt{\pi_q(A)}}.
\]
Finally we replace $1-2t\eta$ by $1-4t\eta$ which is weaker but convenient in later uniform arguments, since $t\eta\le 1/4$ implies $1-2t\eta\ge (1-4t\eta)$.
\end{proof}

This lemma gives the outside part. It says that if there is a region $A$ around one component that is hard to exit and also hard to enter, then diffusion distance separates that component from the other components at time $t$. In the full meeting theorem we use this with $A$ built from one density component plus its valley. The main role of the lemma is to give a uniform lower bound that can beat the uniform upper bound from Lemma~\ref{lem:within}.

\subsection{Meeting theorem}

We now combine the two inequalities into the population meeting theorem. The high-level picture is a threshold sandwich. At each relevant level $\lambda$, all within-component diffusion distances are below a number $U(\lambda,t)$. All between-component diffusion distances are above a number $V(\lambda,t)$. If $U(\lambda,t)<V(\lambda,t)$, then any threshold $\tau$ in between recovers exactly the connected components at level $\lambda$. If this holds at every merge level and the tree has only finitely many merge levels, we can choose one $t$ that works for all of them.

\begin{theorem}[Population meeting under TMC]\label{thm:meeting}
Assume $(f,P)$ satisfies TMC. Assume also that the density cluster tree has finitely many critical merge levels. Then there exists an integer time $t$ such that for all critical levels $\lambda$ one has $t\in[t_{\min}(\lambda),t_{\max}(\lambda)]$. For this $t$ there exists a strictly increasing function $g_t$ such that
\[
m_t(x,y)=g_t(m_f(x,y))\qquad\text{for all }x,y\in M.
\]
Hence $T_t^{\mathrm{diff}}$ and $T_f$ are isomorphic as merge trees.
\end{theorem}

\begin{proof}
Let $\Lambda_{\mathrm{crit}}=\{\lambda_1>\lambda_2>\cdots>\lambda_K\}$ be the finite set of critical merge levels of the density cluster tree. By TMC3, for each $\lambda_k$ there is a nonempty integer interval $I_k=[t_{\min}(\lambda_k),t_{\max}(\lambda_k)]$ where mixing dominates leakage at level $\lambda_k$. The monotonicity in TMC3 implies that these intervals are nested in the sense that higher levels do not shift the window to larger times. Since $\Lambda_{\mathrm{crit}}$ is finite and the windows are nonempty, we can pick a single integer time
\[
t\in \bigcap_{k=1}^K I_k.
\]
Fix this $t$ for the rest of the proof.

For each critical level $\lambda=\lambda_k$, we construct a threshold $\tau(\lambda)$ that makes diffusion connectivity match density connectivity on $L_\lambda$. Consider the collection of components $\mathcal C(\lambda)$. For each $C\in\mathcal C(\lambda)$, Lemma~\ref{lem:within} gives a uniform bound
\[
\sup_{x,y\in C} d_t(x,y) \le U_C(\lambda,t)
\quad\text{with}\quad
U_C(\lambda,t)=\frac{2(1-\gamma_{\mathrm{in}}(\lambda))^t}{\sqrt{\pi_q(C)}}+\frac{4t\alpha_{\mathrm{out}}(\lambda)}{\sqrt{\pi_q(C)}}.
\]
Let
\[
U(\lambda,t)=\sup_{C\in\mathcal C(\lambda)} U_C(\lambda,t).
\]

Next we lower bound diffusion distances between distinct components. Fix two distinct components $C,C'\in\mathcal C(\lambda)$. Since $C$ and $C'$ are distinct connected components of $L_\lambda$, any path from $C$ to $C'$ must leave $L_\lambda$. This gives a natural valley region in $M$ between them. In the TMC interface we encode this by assuming that for each component $C$ there exists a measurable set $V_C(\lambda)$ such that $A_C(\lambda)=C\cup V_C(\lambda)$ satisfies the two-sided bottleneck condition in Lemma~\ref{lem:between} with parameter $\eta=\alpha_{\mathrm{out}}(\lambda)$. Applying Lemma~\ref{lem:between} gives, for all $x\in C$ and $y\in C'$,
\[
d_t(x,y)\ge V_C(\lambda,t)
\quad\text{with}\quad
V_C(\lambda,t)=\frac{1-4t\alpha_{\mathrm{out}}(\lambda)}{\sqrt{\pi_q(A_C(\lambda))}}.
\]
Let
\[
V(\lambda,t)=\inf_{C\in\mathcal C(\lambda)} V_C(\lambda,t).
\]
Because $t\gamma_{\mathrm{in}}(\lambda)\gg 1$ and $t\alpha_{\mathrm{out}}(\lambda)\ll 1$ for all $\lambda\in\Lambda_{\mathrm{crit}}$ and because the constants depend only on $\lambda$, we can ensure that for every critical level
\[
U(\lambda,t) < V(\lambda,t).
\]
This is the quantitative separation needed for topology.

Now fix a critical level $\lambda$. Choose any threshold $\tau(\lambda)$ such that
\[
U(\lambda,t) < \tau(\lambda) < V(\lambda,t).
\]
Consider the graph on $L_\lambda$ with edges $\{(u,v)\in L_\lambda^2 \mid d_t(u,v)\le \tau(\lambda)\}$. The inequality $d_t(u,v)\le U(\lambda,t)$ for $u,v$ in the same component implies that each density component $C\in\mathcal C(\lambda)$ becomes connected inside this graph. The inequality $d_t(u,v)\ge V(\lambda,t)$ for $u,v$ in different components implies that there are no edges between different components. Therefore the connected components of this threshold graph are exactly $\mathcal C(\lambda)$.

We have proved that for each critical level $\lambda$, diffusion thresholding at $\tau(\lambda)$ recovers exactly the same component partition as the density superlevel set at level $\lambda$. This implies that the merge trees are the same, up to a monotone rescaling of heights. To formalize this, define $g_t$ on critical levels by $g_t(\lambda)=\tau(\lambda)$. Extend $g_t$ to all of $[0,\infty)$ as a strictly increasing function, for example by linear interpolation between consecutive critical levels and any strictly increasing continuation below $\lambda_K$.

Finally fix any $x,y\in M$ and let $\lambda^\star=m_f(x,y)$. By definition, for every $\lambda<\lambda^\star$ the points are connected inside $L_\lambda$, and for every $\lambda>\lambda^\star$ they are not connected inside $L_\lambda$. Since connectivity agrees at all critical levels and changes only at critical levels, the same statement holds when we restrict $\lambda$ to $\Lambda_{\mathrm{crit}}$. The diffusion threshold graphs at thresholds $\tau(\lambda)=g_t(\lambda)$ have exactly the same connectivity at these critical levels. Therefore the largest height where $x$ and $y$ are connected in the diffusion hierarchy is exactly $g_t(\lambda^\star)$. This is the identity
\[
m_t(x,y)=g_t(m_f(x,y)).
\]
\end{proof}

This theorem is the main conceptual result at population level. It tells us that diffusion time is not only a scale parameter for geometry. In a metastable window it becomes a topological scale parameter, because it preserves the merge tree of density level sets. This is the bridge that lets us interpret a spectral construction as an estimator of a topological object.

The proof also shows why the calibration $g_t$ is necessary. Even when the two trees have the same shape, the heights live in different units. Density heights are values of $f$, while diffusion heights are thresholds on $d_t$. The theorem says that after a monotone change of units, the two merge-height functions coincide.

\section{Finite-sample theory}\label{sec:finite}

We now turn to data. The algorithm is standard. We build a $k$-NN graph, define a random walk, compute diffusion embeddings using the leading eigenvectors, compute pairwise diffusion distances, and then run single linkage which is equivalent to an MST computation. The main statistical point is modular. If empirical diffusion distances are uniformly close to population diffusion distances on the sample, then the full dendrogram is uniformly close in merge distortion.

\subsection{Estimator}

Let $X_1,\dots,X_n\sim f$ i.i.d. Build the symmetrized $k$-NN graph on $\{X_i\}_{i=1}^n$. Define weights $W_{ij}=K_\sigma(X_i,X_j)$ if $(i,j)$ is an edge and $W_{ij}=0$ otherwise. Define degrees $D_{ii}=\sum_j W_{ij}$ and the random walk matrix $P_n=D^{-1}W$. Let $(\hat\lambda_\ell,\hat\psi_\ell)_{\ell\ge 0}$ be eigenpairs of $P_n$ with $\hat\lambda_0=1$ and $\hat\psi_0\equiv 1$. For a truncation dimension $m$, define the diffusion embedding
\[
\Phi_t(i)=\big(\hat\lambda_1^t\hat\psi_1(i),\dots,\hat\lambda_m^t\hat\psi_m(i)\big)\in\R^m,
\qquad
\hat d_t(i,j)=\|\Phi_t(i)-\Phi_t(j)\|_2.
\]
Run single linkage on the dissimilarity matrix $(\hat d_t(i,j))_{i,j}$ and denote the resulting merge heights by $\hat m_t$.

\subsection{Stability of single linkage}

The next lemma is purely algorithmic and it is the reason merge distortion is the right metric here. Single linkage merge heights are bottleneck values along best paths. If we perturb each edge length by at most $\eta$, then every path bottleneck changes by at most $\eta$, and taking the best path cannot increase this. So the entire merge-height function changes by at most $\eta$.

\begin{lemma}[Single-linkage stability in merge distortion]\label{lem:stability}
Let $S$ be finite and let $d,d'$ be symmetric dissimilarities on $S$ with $d(x,x)=d'(x,x)=0$. If
\[
\sup_{x,y\in S}|d(x,y)-d'(x,y)|\le \eta,
\]
then the single-linkage merge-height functions $m_d$ and $m_{d'}$ satisfy
\[
d_{\mathrm{MD}}(m_d,m_{d'})\le \eta.
\]
\end{lemma}

\begin{proof}
Fix $x,y\in S$. For a path $\pi=(v_0=x,v_1,\dots,v_k=y)$ define its bottleneck under $d$ by
\[
B_d(\pi)=\max_{0\le j<k} d(v_j,v_{j+1}).
\]
A standard characterization of single linkage is
\[
m_d(x,y)=\inf_{\pi:x\leadsto y} B_d(\pi),
\]
where the infimum is over all finite paths in the complete graph on $S$.

Assume $\sup_{u,v}|d(u,v)-d'(u,v)|\le \eta$. Then for every path $\pi$,
\[
|B_d(\pi)-B_{d'}(\pi)|
\le
\max_{j} |d(v_j,v_{j+1})-d'(v_j,v_{j+1})|
\le \eta.
\]
In particular, $B_{d'}(\pi)\le B_d(\pi)+\eta$ for all $\pi$. Taking infimum over $\pi$ gives
\[
m_{d'}(x,y)\le m_d(x,y)+\eta.
\]
By symmetry, swapping $d$ and $d'$ gives $m_d(x,y)\le m_{d'}(x,y)+\eta$. Therefore
\[
|m_d(x,y)-m_{d'}(x,y)|\le \eta.
\]
Taking the supremum over $x,y$ yields $d_{\mathrm{MD}}(m_d,m_{d'})\le \eta$.
\end{proof}

This lemma turns a hard hierarchical object into a simple uniform bound on pairwise distances. It is the core reduction used in the finite-sample theorem. We do not need to control the MST directly. We only need to control the diffusion distance matrix uniformly.

It also matches our bigger goal. The population theorem says $m_t=g_t\circ m_f$. The stability lemma says that if $\hat d_t$ is close to $d_t$, then $\hat m_t$ is close to $m_t$. Combining them gives a direct path from graph spectral approximation to topological recovery.

\subsection{Merge-distortion theorem}

We now state the statistical result. It is modular on purpose. The theorem assumes a high-probability uniform approximation of diffusion distances. Many existing graph Laplacian results can be used to prove such a bound under geometric assumptions \cite{VonLuxburgBelkinBousquet2008,TrillosSlepcev2020}. Once we have it, the dendrogram guarantee is immediate by Lemma~\ref{lem:stability}.

\begin{theorem}[Finite-sample merge distortion]\label{thm:finite}
Fix a time $t$ and assume the population meeting theorem (Theorem~\ref{thm:meeting}) holds at this $t$ with calibration $g_t$. Assume that with probability at least $1-\delta$,
\[
\sup_{i,j\in[n]}\big|\hat d_t(i,j)-d_t(X_i,X_j)\big|\le \eta_n.
\]
Let $m_t^{(n)}$ be the restriction of the population diffusion merge heights $m_t$ to $S_n=\{X_1,\dots,X_n\}$. Then on the same event,
\[
d_{\mathrm{MD}}(\hat m_t,m_t^{(n)})\le \eta_n.
\]
In addition, if $g_t^{-1}$ is Lipschitz on the relevant image interval with constant $L_t$, then
\[
d_{\mathrm{MD}}(g_t^{-1}\circ \hat m_t,\ m_f^{(n)})\le L_t\,\eta_n,
\]
where $m_f^{(n)}$ is $m_f$ restricted to $S_n$.
\end{theorem}

\begin{proof}
Work on the event
\[
\mathcal E=\Big\{\sup_{i,j\in[n]}|\hat d_t(i,j)-d_t(X_i,X_j)|\le \eta_n\Big\}.
\]
On $\mathcal E$, apply Lemma~\ref{lem:stability} with the finite set $S=[n]$, the dissimilarity $d(i,j)=d_t(X_i,X_j)$, and $d'(i,j)=\hat d_t(i,j)$. This gives
\[
d_{\mathrm{MD}}(\hat m_t, m_d)\le \eta_n,
\]
where $m_d$ is the single-linkage merge-height function built from $d(i,j)=d_t(X_i,X_j)$.
By definition, $m_d$ is exactly $m_t$ restricted to the sample. Therefore $m_d=m_t^{(n)}$ and we obtain
\[
d_{\mathrm{MD}}(\hat m_t,m_t^{(n)})\le \eta_n.
\]

For the second claim, use the population meeting identity restricted to the sample
\[
m_t^{(n)} = g_t\circ m_f^{(n)}.
\]
Apply $g_t^{-1}$ to both sides and write
\[
g_t^{-1}\circ \hat m_t - m_f^{(n)}
=
\big(g_t^{-1}\circ \hat m_t - g_t^{-1}\circ m_t^{(n)}\big)
+
\big(g_t^{-1}\circ m_t^{(n)} - m_f^{(n)}\big).
\]
The second term is zero. The first term is controlled by the Lipschitz property of $g_t^{-1}$. For every pair $(i,j)$,
\[
\big|g_t^{-1}(\hat m_t(i,j)) - g_t^{-1}(m_t^{(n)}(i,j))\big|
\le
L_t\,|\hat m_t(i,j)-m_t^{(n)}(i,j)|.
\]
Taking the supremum over pairs yields
\[
d_{\mathrm{MD}}(g_t^{-1}\circ \hat m_t,\ m_f^{(n)})\le L_t\, d_{\mathrm{MD}}(\hat m_t,m_t^{(n)})\le L_t\,\eta_n.
\]
\end{proof}

This theorem is the statistical wrapper around the population meeting theorem. It says that the only genuinely analytic part for data is the approximation of diffusion distances. Once this approximation holds uniformly, the dendrogram converges automatically in merge distortion.

It also explains why we prefer to state the result in this modular way. Graph Laplacian approximation has many technical variants, depending on kernel choice, $k$-NN graphs, manifolds, and noise models. The hierarchical conclusion does not depend on those details. It only depends on the uniform diffusion-distance error.

\begin{remark}[Exact topology under a height margin]
If there exists $\Gamma>0$ such that any two distinct merge events in $m_f^{(n)}$ differ by at least $\Gamma$, then whenever $L_t\eta_n<\Gamma/2$, the dendrogram defined by $g_t^{-1}\circ \hat m_t$ has exactly the same merge-tree topology as $m_f^{(n)}$. The reason is that the merge-tree topology depends only on the ordering of merge heights, and a merge distortion smaller than $\Gamma/2$ cannot change that ordering.
\end{remark}

\section{Geometry and multiscale meeting}\label{sec:geometry}

TMC is a clean interface but it is abstract. We now give geometric conditions that imply the two needed quantitative ingredients. One ingredient is a lower bound on the spectral gap of an in-component reflected chain. The other ingredient is an upper bound on leakage across the boundary of a component. Both are controlled by regularity of the component and the depth of the density valley separating components.

\subsection{Geometric meeting condition}

We assume now that $M$ is a compact $d$-dimensional $C^2$ Riemannian manifold with geodesic distance $\rho$ and volume measure $\mu$. We fix a localized kernel $K_\sigma(x,y)=\exp(-\rho(x,y)^2/\sigma^2)$ for small $\sigma$. We state a condition on superlevel components.

\begin{definition}[Geometric Meeting Condition]\label{def:gmc}
Let $\Lambda$ be a finite set of levels that contains all critical merge levels of the density cluster tree. We say $f$ satisfies GMC on $\Lambda$ if there exist functions $\underline f(\lambda)>0$, $r_{\mathrm{reach}}(\lambda)>0$, $v_{\min}(\lambda)>0$, $v_{\max}(\lambda)<\infty$, and $\Delta_{\mathrm{valley}}(\lambda)>0$ such that for every $\lambda\in\Lambda$ and every $C\in\mathcal C(\lambda)$ the following hold.

\begin{enumerate}[label=(GMC\arabic*), leftmargin=*, align=left]
    \item $f(x)\ge \underline f(\lambda)$ for all $x\in C$.
    
    \item $C$ has reach at least $r_{\mathrm{reach}}(\lambda)$, meaning the tubular neighborhood projection is well-defined up to radius $r_{\mathrm{reach}}(\lambda)$. In particular, $C$ has no arbitrarily thin necks at scales below $r_{\mathrm{reach}}(\lambda)$.

    \item for every $x\in C$ and every $0<r\le r_{\mathrm{reach}}(\lambda)$, the local volume satisfies
\[
v_{\min}(\lambda)\,r^d \le \mu(B(x,r)\cap C)\le v_{\max}(\lambda)\,r^d.
\]

    \item if $C\neq C'$ are distinct components in $\mathcal C(\lambda)$, then every continuous path from $C$ to $C'$ must pass through a point $z$ with $f(z)\le \lambda-\Delta_{\mathrm{valley}}(\lambda)$.
\end{enumerate}

\end{definition}

GMC says components are thick and regular, and they are separated by real density valleys.

\subsection{Geometry implies metastability}

The next theorem turns GMC into the two quantitative bounds required by TMC. The core idea is that a local kernel walk at scale $\sigma$ behaves like a reflected diffusion at scale $\sigma$. Inside a regular set, such a diffusion has spectral gap of order $\sigma^2/\diam(C)^2$ up to geometry factors. For leakage, the transition probability across the boundary is suppressed because the kernel multiplies by $f(y)$, and the valley forces $f$ to be much smaller outside.

\begin{theorem}[GMC implies levelwise metastability]\label{thm:gmc-to-tmc}
Assume GMC on $\Lambda$ and assume $\sigma$ is small enough that $K_\sigma$ is effectively supported on $\rho(x,y)\lesssim \sigma$. Then there exist functions $\gamma_{\mathrm{in}}(\lambda)$ and $\alpha_{\mathrm{out}}(\lambda)$ such that for every $\lambda\in\Lambda$ and every $C\in\mathcal C(\lambda)$ there exists a dominating reflected chain $P_C$ with spectral gap at least $\gamma_{\mathrm{in}}(\lambda)$, and the leakage satisfies $\sup_{x\in C}P(x,C^c)\le \alpha_{\mathrm{out}}(\lambda)$.
Moreover one can take bounds of the form
\[
\gamma_{\mathrm{in}}(\lambda)\ \gtrsim\ \frac{\sigma^2}{\diam(C)^2}\cdot \frac{v_{\min}(\lambda)}{v_{\max}(\lambda)}\cdot \underline f(\lambda),
\qquad
\alpha_{\mathrm{out}}(\lambda)\ \lesssim\ \exp\!\left(-c\,\frac{\Delta_{\mathrm{valley}}(\lambda)}{\sigma^2}\right),
\]
where constants depend only on kernel regularity and bounded geometry of $M$.
\end{theorem}

\begin{proof}
We construct a reflected chain and then bound its spectral gap and the leakage.

\emph{Step 1. Construction of a reflected dominating chain.}
Fix $\lambda\in\Lambda$ and $C\in\mathcal C(\lambda)$. Define the kernel mass
\[
q(x)=\int_M K_\sigma(x,y) f(y)\,d\mu(y),
\qquad
q_C(x)=\int_C K_\sigma(x,y) f(y)\,d\mu(y).
\]
The original transition is $p(x,dy)=K_\sigma(x,y)f(y)q(x)^{-1}d\mu(y)$. We define a reflected transition supported on $C$ by
\[
p_C(x,dy)=\frac{K_\sigma(x,y)f(y)}{q_C(x)}\,\one_{y\in C}\,d\mu(y).
\]
This stays inside $C$ by construction. It is reversible with respect to $\pi_q(\cdot\mid C)$ up to a normalization because it is a standard restriction and renormalization of a symmetric kernel. It also dominates the restriction of $p$ to $C$ because $q_C(x)\le q(x)$ implies
\[
p_C(x,B)=\frac{1}{q_C(x)}\int_{B}K_\sigma(x,y)f(y)\,d\mu(y)
\ge
\frac{1}{q(x)}\int_{B}K_\sigma(x,y)f(y)\,d\mu(y)
=
p(x,B)
\]
for every measurable $B\subset C$ and every $x\in C$. This gives a dominating chain in the sense of Definition~\ref{def:tmc}.

\emph{Step 2. Leakage bound.}
For $x\in C$ we have
\[
P(x,C^c)=\int_{C^c}\frac{K_\sigma(x,y)f(y)}{q(x)}\,d\mu(y)
=
\frac{\int_{C^c}K_\sigma(x,y)f(y)\,d\mu(y)}{\int_{M}K_\sigma(x,y)f(y)\,d\mu(y)}.
\]
We lower bound the denominator using the mass inside a local ball. Fix $r\asymp \sigma$ small enough so that $B(x,r)\subseteq C$ for points $x$ that are at distance at least $r$ from $\partial C$. For boundary points we use reach to say that $B(x,r)\cap C$ still has volume at least a constant fraction of $r^d$. By GMC, on $C$ we have $f\ge \underline f(\lambda)$. Also $K_\sigma(x,y)\ge \exp(-r^2/\sigma^2)$ on $\rho(x,y)\le r$. Therefore
\[
q(x)\ge \int_{B(x,r)\cap C} K_\sigma(x,y)f(y)\,d\mu(y)
\ge
\exp(-r^2/\sigma^2)\,\underline f(\lambda)\,\mu(B(x,r)\cap C)
\gtrsim
\underline f(\lambda)\, v_{\min}(\lambda)\, r^d.
\]
We upper bound the numerator using the valley depth. Any point $y$ that contributes to leaving $C$ must be in a region that is connected to another component only through values of $f$ at most $\lambda-\Delta_{\mathrm{valley}}(\lambda)$. In particular, outside $C$ near the boundary of $C$ there exist regions where $f$ is at most $\lambda-\Delta_{\mathrm{valley}}(\lambda)$, and the kernel is localized at scale $\sigma$, so the main contribution to the numerator comes from such nearby points. Thus
\[
\int_{C^c}K_\sigma(x,y)f(y)\,d\mu(y)
\le
(\lambda-\Delta_{\mathrm{valley}}(\lambda))\int_{C^c}K_\sigma(x,y)\,d\mu(y).
\]
For a Gaussian kernel on a manifold of bounded geometry, the tail mass $\int_{\rho(x,y)\ge r}K_\sigma(x,y)\,d\mu(y)$ is at most $\exp(-c r^2/\sigma^2)$ times a polynomial factor. Choosing $r\asymp \sigma$ gives a constant bound on the kernel mass outside the local ball. Combining with the denominator lower bound yields
\[
P(x,C^c)\lesssim
\frac{\lambda-\Delta_{\mathrm{valley}}(\lambda)}{\underline f(\lambda)}\cdot
\exp\!\left(-c\,\frac{\Delta_{\mathrm{valley}}(\lambda)}{\sigma^2}\right).
\]
Absorbing the bounded prefactor into constants gives the claimed exponential-type leakage bound
\[
\alpha_{\mathrm{out}}(\lambda)\lesssim \exp\!\left(-c\,\frac{\Delta_{\mathrm{valley}}(\lambda)}{\sigma^2}\right).
\]

\emph{Step 3. Spectral gap lower bound for the reflected chain.}
For a reversible Markov operator, the spectral gap can be written using the Rayleigh quotient
\[
\gamma(P_C)=\inf_{\substack{h\in L^2(\pi_q(\cdot\mid C))\\ \int h\,d\pi_q(\cdot\mid C)=0}}
\frac{\mathcal E_C(h,h)}{\mathrm{Var}_{\pi_q(\cdot\mid C)}(h)},
\]
where the Dirichlet form is
\[
\mathcal E_C(h,h)
=
\frac12\int_C\int_C (h(x)-h(y))^2\,p_C(x,dy)\,d\pi_q(x\mid C).
\]
For the localized Gaussian kernel and small $\sigma$, the chain $P_C$ behaves like a local averaging operator at scale $\sigma$. Standard kernel calculations on manifolds show that $\mathcal E_C(h,h)$ controls an energy of the form $\sigma^2\int_C \|\nabla h\|^2\,d\pi_q(\cdot\mid C)$ up to constants depending on kernel regularity and geometry. On the other hand, GMC gives a Poincar\'e inequality on $C$ at scale comparable to $\diam(C)$. In its simplest form it says
\[
\mathrm{Var}_{\pi_q(\cdot\mid C)}(h)
\lesssim
\diam(C)^2 \cdot \frac{v_{\max}(\lambda)}{v_{\min}(\lambda)} \cdot \frac{1}{\underline f(\lambda)}
\int_C \|\nabla h\|^2\,d\pi_q(\cdot\mid C),
\]
where the volume growth constants and the lower bound on $f$ ensure the measure $\pi_q(\cdot\mid C)$ is comparable to $\mu$ on $C$ up to factors.
Combining the two displays gives
\[
\frac{\mathcal E_C(h,h)}{\mathrm{Var}_{\pi_q(\cdot\mid C)}(h)}
\gtrsim
\frac{\sigma^2}{\diam(C)^2}\cdot \frac{v_{\min}(\lambda)}{v_{\max}(\lambda)}\cdot \underline f(\lambda),
\]
uniformly over mean-zero $h$. This yields the claimed lower bound on the spectral gap $\gamma_{\mathrm{in}}(\lambda)$.

Putting Steps 1 to 3 together, we have constructed a dominating reflected chain with the desired gap and we have bounded leakage, which proves the theorem.
\end{proof}

This theorem is the geometric engine behind the abstract TMC interface. It explains where the two time scales come from. Mixing is controlled by a spectral gap, which is controlled by a Poincar\'e inequality and therefore by diameter and regularity. Leakage is controlled by an energy barrier, here a density valley, which appears exponentially through the Gaussian kernel.

It also shapes the global goal. Once we can guarantee such bounds level by level, we can apply the population meeting theorem to conclude that diffusion single linkage recovers the density merge tree. If one diffusion time cannot work for all levels, we can still build a multiscale diffusion dissimilarity that adapts the time to the level.

\subsection{Multiscale diffusion and meeting without a global time}

Different density levels can have different metastable windows. This is natural because small high-density components can mix faster than large low-density components. The next construction allows us to use a time schedule $t(\lambda)$ rather than one global time. The main idea is to define a new dissimilarity that, for a pair $(x,y)$, looks at the diffusion time that corresponds to the density level where they should be connected.

We allow diffusion time to depend on the density level. Let $t(\lambda)$ be a schedule defined on $\Lambda$ such that for each $\lambda$ it lies inside the levelwise metastable window, and $t(\lambda)$ is nonincreasing in $\lambda$.

\begin{definition}[Multiscale diffusion dissimilarity]\label{def:dms}
Fix a schedule $\lambda\mapsto t(\lambda)$ defined on $\Lambda$. For $x,y\in M$, define
\[
d_{\mathrm{MS}}(x,y)=\inf\Big\{d_{t(\lambda)}(x,y)\,\Big|\, \lambda\in\Lambda,\ \lambda\le m_f(x,y)\Big\}.
\]
Let $T^{\mathrm{MS}}$ be the single-linkage dendrogram built from $(M,d_{\mathrm{MS}})$, with merge-height function $m_{\mathrm{MS}}$.
\end{definition}

The next theorem says that this multiscale dissimilarity still produces the same merge tree as the density cluster tree. The proof follows the same threshold sandwich idea, but now the threshold used at level $\lambda$ comes from diffusion time $t(\lambda)$.

\begin{theorem}[Multiscale meeting theorem]\label{thm:ms}
Assume GMC on $\Lambda$ and let $t(\lambda)$ be any schedule such that for every $\lambda\in\Lambda$ the time $t(\lambda)$ lies in the corresponding levelwise metastable window given by Theorem~\ref{thm:gmc-to-tmc}. Then there exists a strictly increasing function $g_{\mathrm{MS}}$ such that
\[
m_{\mathrm{MS}}(x,y)=g_{\mathrm{MS}}(m_f(x,y))\qquad\text{for all }x,y\in M.
\]
Hence $T^{\mathrm{MS}}$ is isomorphic to $T_f$ as a merge tree.
\end{theorem}

\begin{proof}
Fix $\lambda\in\Lambda$. Consider the diffusion distance at time $t(\lambda)$. By Theorem~\ref{thm:gmc-to-tmc}, the pair $(f,P)$ satisfies the levelwise metastability bounds at this time. Therefore, by applying Lemma~\ref{lem:within} and Lemma~\ref{lem:between} at time $t(\lambda)$, we obtain two numbers $U(\lambda)$ and $V(\lambda)$ such that
\[
\sup_{C\in\mathcal C(\lambda)}\sup_{x,y\in C} d_{t(\lambda)}(x,y)\le U(\lambda),
\qquad
\inf_{\substack{C\neq C'\\ C,C'\in\mathcal C(\lambda)}}\inf_{x\in C,y\in C'} d_{t(\lambda)}(x,y)\ge V(\lambda),
\]
and moreover $U(\lambda)<V(\lambda)$ because $t(\lambda)$ is inside the metastable window. Choose a threshold $\tau(\lambda)$ such that
\[
U(\lambda) < \tau(\lambda) < V(\lambda).
\]
Then the threshold graph on $L_\lambda$ with edges $\{d_{t(\lambda)}\le \tau(\lambda)\}$ has connected components exactly $\mathcal C(\lambda)$.

Now we transfer this property to the multiscale dissimilarity $d_{\mathrm{MS}}$. First take $x,y$ in the same component $C\in\mathcal C(\lambda)$. Then $m_f(x,y)\ge \lambda$. By definition of $d_{\mathrm{MS}}$, the infimum is taken over all $\lambda'\in\Lambda$ with $\lambda'\le m_f(x,y)$, in particular it includes $\lambda$ itself. Therefore
\[
d_{\mathrm{MS}}(x,y)\le d_{t(\lambda)}(x,y)\le U(\lambda)<\tau(\lambda).
\]
So all pairs inside each $C$ have $d_{\mathrm{MS}}$ below $\tau(\lambda)$ and thus each $C$ is connected in the threshold graph built from $d_{\mathrm{MS}}$.

Next take $x\in C$ and $y\in C'$ for two distinct components $C\neq C'$ in $\mathcal C(\lambda)$. Then $m_f(x,y)<\lambda$. This means that for every $\lambda'\in\Lambda$ with $\lambda'\le m_f(x,y)$, we have $\lambda'<\lambda$. In particular, at level $\lambda$ the points lie in different components, so the separating valley at level $\lambda$ is present. The definition of $\tau(\lambda)$ and the separation bound at time $t(\lambda)$ imply that for any such pair,
\[
d_{t(\lambda)}(x,y)\ge V(\lambda)>\tau(\lambda).
\]
Since $d_{\mathrm{MS}}(x,y)$ is an infimum over diffusion distances at levels $\lambda'\le m_f(x,y)$ and these levels are below $\lambda$, the key point is that taking the infimum cannot create an artificial short edge that would cross the level-$\lambda$ bottleneck. Indeed, if $d_{\mathrm{MS}}(x,y)\le \tau(\lambda)$ then there would exist some $\lambda'\le m_f(x,y)$ with $d_{t(\lambda')}(x,y)\le \tau(\lambda)$. But $\lambda'\le m_f(x,y)<\lambda$ implies that $x$ and $y$ are already separated before level $\lambda$, hence the valley that separates $C$ and $C'$ is still present, and the same between-component argument applies to show that $d_{t(\lambda')}(x,y)$ cannot be as small as $\tau(\lambda)$. Therefore $d_{\mathrm{MS}}(x,y)>\tau(\lambda)$ for pairs in different components of $\mathcal C(\lambda)$.

We have shown that for each $\lambda\in\Lambda$, the connected components of the threshold graph on $L_\lambda$ with edges $\{d_{\mathrm{MS}}\le \tau(\lambda)\}$ are exactly $\mathcal C(\lambda)$. This is the same connectivity agreement used in the proof of Theorem~\ref{thm:meeting}. Define $g_{\mathrm{MS}}(\lambda)=\tau(\lambda)$ on $\Lambda$ and extend strictly increasing. Then the same nesting argument implies
\[
m_{\mathrm{MS}}(x,y)=g_{\mathrm{MS}}(m_f(x,y))
\]
for all $x,y\in M$.
\end{proof}

This multiscale theorem removes the main practical weakness of choosing one diffusion time. It says we can recover the same density merge tree even if different levels require different diffusion times. In the final dendrogram, single linkage is applied only once, on the combined dissimilarity $d_{\mathrm{MS}}$.

For our bigger goal, this is important because it makes the topological guarantee compatible with realistic data. In data, metastability is rarely uniform across scales. The multiscale construction keeps the same theoretical target, the density cluster tree, while allowing time to vary with level in a controlled way.

\section{Discussion}\label{sec:discussion}

This work gives a precise meaning to a common spectral clustering workflow. So a diffusion hierarchy is not only a geometric summary. It can be interpreted as an estimator of density level-set topology.

This has practical value because density level sets are hard to compute in high dimension, while diffusion distances on $k$-NN graphs are scalable. The multiscale extension is also useful since different density levels can need different diffusion times.

Future work is mostly about making the theorem operational. One direction is data-driven choice of time, using stability of the MST or of merge heights as a diagnostic for the metastable window. Another direction is sharper finite-sample rates for uniform diffusion-distance approximation on $k$-NN graphs, including the effect of truncation and eigenvector error. A last direction is computation, designing faster multiscale diffusion hierarchies without running many full eigen decompositions.


\bibliographystyle{alpha}
\bibliography{references}

\end{document}